\documentclass[sigconf, nonacm]{acmart} %
\usepackage{multirow}
\usepackage{makecell}
\usepackage{subfloat}
\usepackage{placeins}

\begin{document}

\title{\texttt{ClimDetect}: A Benchmark Dataset for Climate Change Detection and Attribution}

\author{Sungduk Yu}
\authornote{Corresponding author: sungduk.yu@intel.com}
\affiliation{%
  \institution{Intel Labs}
  \city{Santa Clara}
  \state{CA}
  \country{USA}
}

\author{Brian L. White}
\affiliation{%
  \institution{UNC Chapel Hill}
  \city{Chapel Hill}
  \state{NC}
  \country{USA}
}

\author{Anahita Bhiwandiwalla}
\affiliation{%
  \institution{Intel Labs}
  \city{Santa Clara}
  \state{CA}
  \country{USA}
}

\author{Musashi Hinck}
\affiliation{%
  \institution{Intel Labs}
  \city{Santa Clara}
  \state{CA}
  \country{USA}
}

\author{Matthew Lyle Olson}
\affiliation{%
  \institution{Intel Labs}
  \city{Santa Clara}
  \state{CA}
  \country{USA}
}

\author{Yaniv Gurwicz}
\affiliation{%
  \institution{Intel Labs}
  \city{Santa Clara}
  \state{CA}
  \country{USA}
}

\author{Raanan Y. Rohekar}
\affiliation{%
  \institution{Intel Labs}
  \city{Santa Clara}
  \state{CA}
  \country{USA}
}

\author{Tung Nguyen}
\affiliation{%
  \institution{UCLA}
  \city{Los Angeles}
  \state{CA}
  \country{USA}
}

\author{Vasudev Lal}
\affiliation{%
  \institution{Intel Labs}
  \city{Santa Clara}
  \state{CA}
  \country{USA}
}

\renewcommand{\shortauthors}{Yu et al. (2025)}

\begin{abstract}
Detecting and attributing temperature increases driven by climate change is crucial for understanding global warming and informing adaptation strategies. However, distinguishing human-induced climate signals from natural variability remains challenging for traditional detection and attribution (D\&A) methods, which rely on identifying specific "fingerprints"—spatial patterns expected to emerge from external forcings such as greenhouse gas emissions. Deep learning offers promise in discerning these complex patterns within expansive spatial datasets, yet the lack of standardized protocols has hindered consistent comparisons across studies. 

To address this gap, we introduce \texttt{ClimDetect}, a standardized dataset comprising 1.17M daily climate snapshots paired with target climate change indicator variables. The dataset is curated from both CMIP6 climate model simulations and real-world observation-assimilated reanalysis datasets (ERA5, JRA-3Q, and MERRA-2), and is designed to enhance model accuracy in detecting climate change signals. \texttt{ClimDetect} integrates various input and target variables used in previous research, ensuring comparability and consistency across studies. We also explore the application of vision transformers (ViT) to climate data—a novel approach that, to our knowledge, has not been attempted before for climate change detection tasks. Our open-access data serve as a benchmark for advancing climate science by enabling end-to-end model development and evaluation. ClimDetect is publicly accessible via Hugging Face dataset repository at: \url{https://huggingface.co/datasets/ClimDetect/ClimDetect}.
\end{abstract}


\maketitle

\begin{figure*}
    \centering
    \includegraphics[width=1\linewidth]{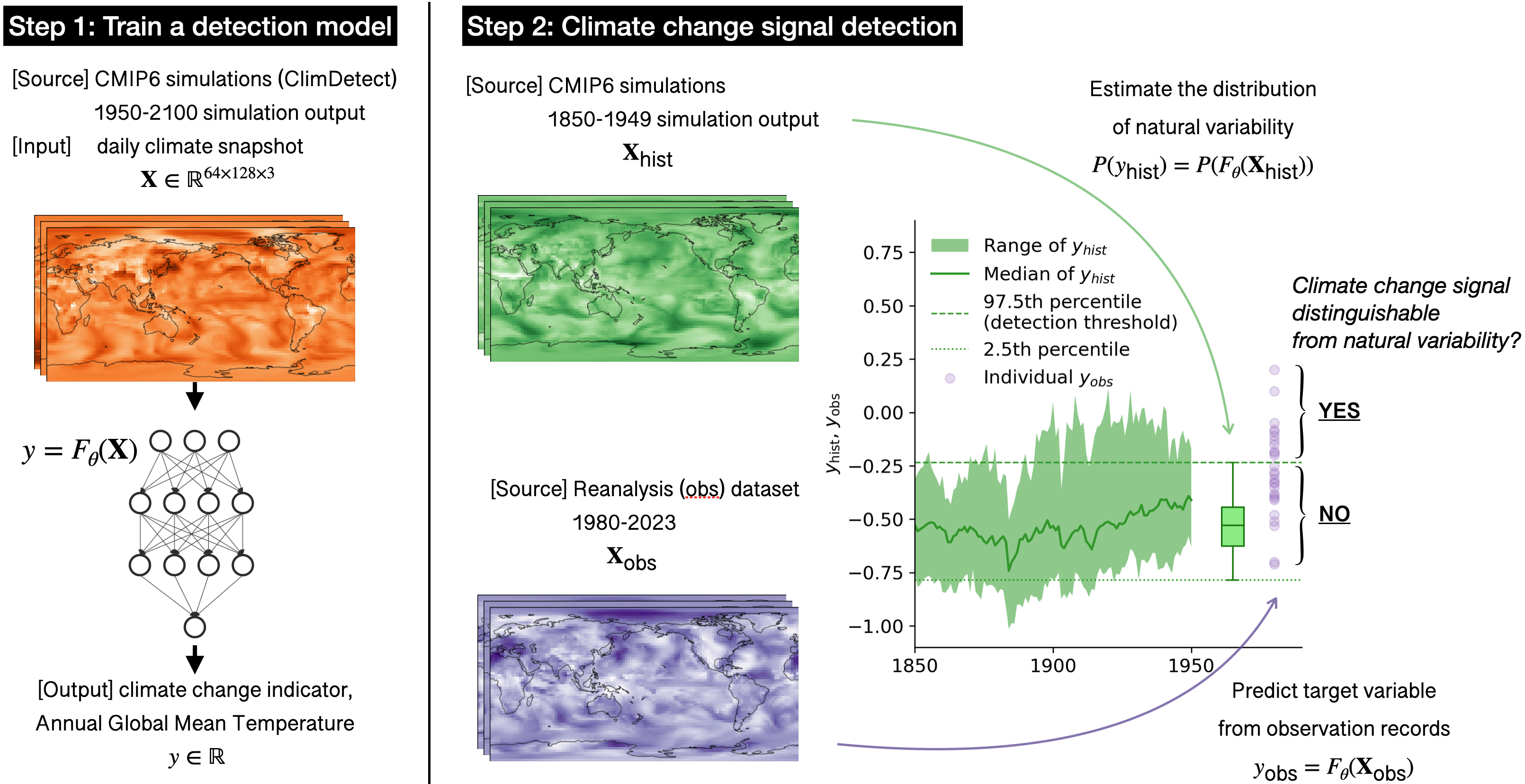}
    \caption{Overview of the machine learning pipeline for climate change detection and attribution using the \texttt{ClimDetect} dataset. The diagram illustrates the workflow from input daily climate model variables (surface air temperature, humidity, precipitation), through a neural network model, to the target annual global mean temperature (AGMT). The diagram features climate field maps distinguished by color to denote independent datasets: the training dataset in orange, the historical (i.e., pre-warming) dataset in green, and the observation dataset in purple. $F_\theta$ denotes a detection model (e.g., vision transformer, CNN, etc.), where $\theta$ represents the parameters of the model. One purple dot represent an individual estimates from a single observation sample. For detailed information, see Section \ref{sec:detection_method}}
    \label{fig:cartoon1}
\end{figure*}

\section{Introduction}
Climate change, particularly the increase in global temperature in response to anthropogenic greenhouse gas emissions, has emerged as one of the most pressing environmental challenges of the 21st century. The Intergovernmental Panel on Climate Change (IPCC) has highlighted the importance of understanding the drivers of these changes in order to implement effective mitigation and adaptation strategies \citep{ipcc2021wg1,ipcc_ar6_wg2,ipcc_ar6_wg3}. A key challenge in climate science has been developing methodologies for the detection and attribution (D\&A) of climate signals, in order to differentiate the subtly emerging, long-term signals of human-induced warming from the inherently more volatile and transient patterns of natural climate variability \cite{Schneider.1994,Deser.2012hl3e,Hawkins.2012}.

The concept of D\&A in climate science has focused on identifying the 'fingerprint' of climate change—a unique spatial pattern in climate response variables that can be attributed to specific external forcings, such as increased concentrations of greenhouse gases. The IPCC's Sixth Assessment Report (AR6) \citep{ipcc2021wg1} emphasizes the advancements in D\&A methodologies that have bolstered confidence in the attribution of observed climate change to human activities. Despite these advancements, significant challenges persist, including the need for standardizing D\&A methodologies across climate forcing metrics and response variables, and improving D\&A sensitivity, in order to detect climate change signals in short-term variability and extremes.

Traditional approaches in climate D\&A have leveraged statistical methods to discern climate change "fingerprints" (that is, patterns in climate response variables that can be attributed to external forcing, such as greenhouse gas emissions). Many studies have used principal component analysis (PCA; also known as empirical orthogonal function, EOF, analysis in the climate community) to derive spatial fingerprints from long-term climate ensembles \citep{Santer.1994, Santer.2007, Santer.2018}. However, recent studies have advanced the state of the art, for example by showing it is possible to detect the climate change signal in daily snapshots of global weather \citep{Sippel.2020, Sippel.2021}, and by leveraging the advantages of machine learning for pattern recognition \cite[e.g.,][]{Barnes.2019, Barnes.2020, Rader.2022, Ham.2023} to learn spatial fingerprints and forced climate response directly from large climate datasets. 
These recent result show that deep learning can provide useful tools for D\&A, due to the ability of models to uncover intricate patterns in large climate datasets. However, the application of sophisticated models requires large, diverse and balanced dataset that encapsulates a wide range of climate response variables, forcing metrics, natural variability, and climate models. Nevertheless, a dedicated dataset for climate D\&A tasks is not yet available; most existing climate-related datasets focus on weather prediction or climate emulation (see Section \ref{sec:related} for details).

In response to these needs, we introduce \texttt{ClimDetect}, a comprehensive dataset designed to promote and standardize \textit{data-driven} climate change detection tasks (Figure \ref{fig:cartoon1}). This dataset includes 1,173,913 daily climate snapshots---paired with target climate change indicator variables---of historical and future climate scenarios from both the Coupled Model Intercomparison Project Phase 6 (CMIP6) model ensemble \citep{cmip6_overview} and also from three popular reanalysis datasets (ERA5 \citep{sociera5}, JRA-3Q \citep{kosaka2024jra}, and MERRA-2 \citep{gelaro2017modern}), carefully curated by subject matter experts to foster the development of models capable of detecting climate change signals in daily weather patterns. \texttt{ClimDetect} aims to address the fragmentation in previous studies by standardizing the input and target variables used in climate fingerprinting, promoting consistency and comparability across D\&A research efforts.
Our research extends current understanding beyond traditional methodologies by incorporating modern machine learning architectures, i.e. Vision Transformers (ViT) \citep{dosovitskiy2020image}, into climate science. ViTs have shown amazing performance on natural image tasks and are adapted in our study to analyze spatial climate data. 

By providing open access to the \texttt{ClimDetect} dataset, our work sets a benchmark for future studies, encouraging the exploration of diverse modeling techniques in the climate science community. 
With this work, we hope to foster scientific research and addresses societal goals by deepening understanding and mitigation of climate change impacts.

\section{Related Work\protect\footnote{For readers unfamiliar with climate-science concepts and terminology, please see the brief overview in Appendix \ref{sec:appendix_concepts}, which introduces key background relevant to this section.}}
\label{sec:related}

\subsection{Climate Detection and Attribution Studies}

Previous D\&A work in the climate science community has focused on distinguishing the climate signal from internal variability by finding spatial fingerprints. However, methodologies differ in (i) the statistical or modeling approach used to discover the fingerprint, (ii) the climate time scales of focus, (iii) the climate response variables under study, and (iv) the target metric chosen to represent the climate forcing.

Santer et al. \citep{Santer.1994} use principal component analysis (PCA) analysis to investigate the fingerprints of climate variables, including water vapor \citep{Santer.2007} and tropospheric temperature \citep{Santer.2018}. They project observations onto leading PCA modes from climate model ensembles to show statistical significance compared to control runs without greenhouse gas forcing, primarily examining multi-decadal periods.

Sippel et al. detect climate fingerprints in daily weather variables (surface temperature and humidity) via regularized linear (ridge) regression \citep{Sippel.2020} and anchor regression \citep{Sippel.2021} on large CMIP5/6 ensembles \citep{taylor2012overview,cmip6_overview}, using annual global mean temperature (AGMT) as the target metric. They project the daily variables (with or without mean removal) onto these fingerprints to predict current observations, then compare the predictions to a pre-warming (1850–1950) climate baseline.

Barnes and co-authors \citep{Barnes.2019, Barnes.2020} apply machine learning to find the spatial pattern of warming, treating it as a classification task. They train an MLP on a large CMIP5 ensemble with annual temperature and precipitation inputs to predict the model year \citep{Barnes.2019, Barnes.2020}. Detectability is defined as the "time of emergence" relative to a control period (1920--1959) \citep{Rader.2022}, and interpretable ML techniques (layer-wise relevance propagation \citep[e.g.,][]{Barnes.2020, Labe.2021, Rader.2022} and backward optimization \citep{Barnes.2020, Toms.2020}) visualize the learned spatial patterns. Ham et al. \citep{Ham.2023} use a CNN to predict AGMT using only precipitation—a weaker signal than temperature—yet show potential for deep learning to yield more sensitive D\&A methods.

These prior works underscore the value of standardized datasets and methodologies for improving and comparing D\&A approaches. We develop \texttt{ClimDetect} to address these needs by creating a balanced and diverse dataset specifically designed for D\&A research.

\subsection{Climate Datasets for ML}
A handful of previous works have created large climate and weather datasets for training and benchmarking deep learning models. Datasets like WeatherBench \cite{Rasp.2020}, WeatherBench2 \cite{Rasp.2023}, and ChaosBench \cite{Nathaniel.2024} have advanced ML-based weather prediction by creating consistent benchmarks for forecasting at different lead times. This shift has enabled data-driven forecasts \cite[e.g.,][]{Pathak.2022,Lam.2023, Bi.2023, Nguyen.2023rm} to approach parity with traditional numerical weather prediction (NWP) models.

For climate-focused tasks, benchmark datasets have also emerged. ClimSim Yu et al. \citep{Yu.2023ht}, the largest dataset for hybrid physics–ML modeling, enables improved convective parameterization in climate models. ClimateBench \citep{Watson‐Parris.2022} is an ML-ready dataset for climate emulation, consisting of forcing variables (CO$_2$, CH$_4$, SO$_2$) and outputs from a single CMIP6 model over multiple scenarios, enabling the prediction of future annual-mean climate variables. 
Kaltenborn ett al. \citep{Kaltenborn.2023} created ClimateSet, an extension to 36 CMIP6 models with monthly forcing and output variables for historical and future scenarios. 
Nguyen et al. \citep{Nguyen.20230u} developed ClimateLearn, an open-source PyTorch library for training and evaluating ML models in both weather and climate contexts.

While these datasets have significantly advanced weather forecasting and climate emulation, they do not directly address the detection and attribution (D\&A) of anthropogenic climate signals. This gap motivates our work, which introduces \texttt{ClimDetect}, a dedicated benchmark dataset for D\&A tasks.

\section{\texttt{ClimDetect} Dataset}
\label{sec:data}

We develop \texttt{ClimDetect}, a dataset comprising 1,173,913 daily climate snapshots from the CMIP6 model ensemble and 74,825 daily snapshots from reanalysis data to enable detection and attribution (D\&A) studies. The dataset pairs daily snapshots of key climate variables (inputs) with a climate indicator variable representing climate forcing (target). In essence, \texttt{ClimDetect} is designed to detect climate change signals in the input data, with the target variable serving as a proxy for climate change.

\subsection{Variables}
\textbf{Input variables}. For input variables, we selected three key climate variables: surface 2-meter air temperature (tas), surface 2-meter specific humidity (huss), and total precipitation (pr) (see Table \ref{tab:variables}). These variables were chosen because they are widely recognized as important climate response indicators and have been extensively studied in previous detection and attribution research \citep[e.g.,][]{Barnes.2019,Barnes.2020,Sippel.2020,Sippel.2021,Ham.2023}. This ensures that our dataset aligns with established scientific methodologies and promotes comparable, replicable research. Each input sample, \( \mathbf{X} \), is a 3-dimensional matrix with dimensions \( 3 \times 64 \times 128 \), where 64 and 128 correspond to the spatial grid points (latitude and longitude), and 3 represents the three input climate variables---that is, \( \mathbf{X} \in \mathbb{R}^{64 \times 128 \times 3} \). This can be thought of analogously as an RGB image with 64 by 128 pixels.

\textbf{Output variables}. Our primary target variable is the annual global mean temperature (AGMT), defined as the annual mean of spatially-averaged surface air temperature. Also known as global surface air temperature (GSAT), AGMT is a widely used proxy for climate change \citep{allen2018special, ipcc2021wg1} and is central to many climate studies. In addition, we include the “year” as a secondary target variable. This is justified because in a warming climate, global temperatures generally rise with each passing year. Although this variable is not used in our benchmark evaluations, it is provided for completeness, as several D\&A studies have used the year as an alternative proxy for climate change \citep[e.g.,][]{Barnes.2019,Barnes.2020,Rader.2022}. Target variable, \( y \), is a scalar, that is, \( y \in \mathbb{R} \).

\begin{table}
\caption{List of input and target variables in the \texttt{ClimDetect} dataset.}
\begin{tabular}{ll}
\toprule
Variable                                & Size  \\
\midrule
\underline{Input}:&\\
Surface 2-meter temperature (tas)        & {(}1, 64, 128{)} \\
Surface 2-meter specific humidity (huss) & {(}1, 64, 128{)} \\
Total precipitation (pr)                 & {(}1, 64, 128{)} \\
\underline{Target}:&\\
Annual global mean temperature (AGMT) & (1)     \\
Year                                  & (1)     \\
\bottomrule
\end{tabular}
\label{tab:variables}
\end{table}

\subsection{Data Source}
To train our climate change detection models, we used global climate model outputs from CMIP6 archive. For evaluation on earth observation records, we utilized three modern reanalysis products.

\textbf{CMIP6}.
CMIP6 is a globally coordinated climate model experiment initiative, featuring over 100 models from more than 50 research groups \cite{cmip6_overview}. It encompasses historical simulations covering 1850 to 2014, along with ScenarioMIP projections extending from 2015 to 2100 under diverse socioeconomic pathways, thereby offering a comprehensive dataset for climate research (See Appendix \ref{sec:appendix_concepts} for further details). CMIP6 data is publicly available via the Earth System Grid Federation (ESGF) and other redistribution platforms\footnote{\url{https://wcrp-cmip.org/cmip-data-access/}}.

\textbf{Reanalysis}.
Atmospheric reanalysis datasets synthesize observations with weather forecast model outputs to create continuous, globally comprehensive climate records—offering a more consistent alternative to sparse observational data alone. In other words, a reanalysis dataset represents an optimal integration of model outputs and observational data, effectively accounting for both model errors and observation measurement errors. We use three modern reanalysis datasets that serve as the closest alternative to direct observations when continuous data coverage is required for robust model evaluation and hypothesis testing: ERA5 (data span: 1940–2024) \cite{sociera5}, JRA-3Q (1950–2024) \cite{kosaka2024jra}, and MERRA-2 (1980–2024) \cite{gelaro2017modern}. Note that although these datasets are anchored in observational data during overlapping periods, their values differ due to variations in the underlying weather forecast models, data assimilation algorithms, and the observational data incorporated. These datasets are also publicly available through their official websites\footnote{(ERA5) \url{https://www.ecmwf.int/en/forecasts/dataset/ecmwf-reanalysis-v5/}\\ (MERRA2) \url{https://gmao.gsfc.nasa.gov/reanalysis/merra-2/data\_access/}\\ (JRA-3Q) \url{https://jra.kishou.go.jp/JRA-3Q/index\_en.html}}
and other redistribution platforms. In the context of our study, `observations' refers to reanalysis datasets. 

\subsection{Data Collection}
\label{sec:datacollection}

Given that participation in CMIP6 is voluntary, the availability of simulated climate variables and the number of ensemble simulations differ across the various climate models. To ensure consistency and reliability in our dataset, we implemented a three of criteria for model selection.

\textbf{Model Requirements:} Each model must include simulations from the historical experiment covering the entire simulation period from 1850 to 2014, as well as from at least one of the selected ScenarioMIP experiments (SSP2-4.5 and SSP3-7.0) for the period from 2015 to 2100.

\textbf{Variable Availability:} It is essential that all three key climate variables---surface air temperature, surface air humidity, and total precipitation rate---are available for each simulation. This criterion guarantees that our dataset consistently represents the primary factors influencing global climate patterns.

\textbf{Temporal Coverage:} The models selected must provide data for the entire duration of the specified experiments, ensuring comprehensive coverage and facilitating accurate long-term climate analysis.

By adhering to these stringent selection criteria, we aim to maximize the robustness and applicability of the \texttt{ClimDetect} dataset, enabling detailed analysis and modeling of climate dynamics under various emission scenarios as projected by CMIP6. The details of the selected data are included in Appendix \ref{App:cmip6_info}. All CMIP6 data was accessed from the Registry of Open Data on AWS \footnote{https://registry.opendata.aws/cmip6/} available under CC BY 4.0 License. 

\subsection{Postprocessing}
\texttt{ClimDetect} is designed to be a machine learning (ML)-ready dataset and has therefore undergone specific postprocessing steps to standardize the data for optimal ML model performance. The processing of input variables follows a method similar to z-score standardization, tailored to address the unique characteristics of climate data. These postprocessing steps are commonly employed in previous studies \citep[e.g.,][]{Sippel.2020,Rader.2022,Ham.2023}, and we adopt them here to maintain consistency with established methodologies. The postprocessing involves two primary steps:

\textbf{Removal of the Climatological Daily Seasonal Cycle}. Each input sample \( \mathbf{X} \), which is represented as a three-dimensional array with dimensions [channel, latitude, longitude], is adjusted to remove the climatological daily seasonal cycle, denoted as \( \mathbf{X}_{\text{clim}} \). This step encourages analysis on anomalies rather than absolute values, which can be heavily influenced by seasonal effects that often overshadow more subtle interannual or long-term variations in AGMT. Mathematically, the anomaly \( \mathbf{X}' \) for each data point is calculated as:
   $ \mathbf{X}' = \mathbf{X} - \mathbf{X}_{\text{clim}} $
   where \( \mathbf{X}_{\text{clim}} \) represents the long-term average for that particular day at each channel, latitude, and longitude point, effectively normalizing the data across years to highlight deviations from typical patterns.

\textbf{Standardization of Anomalies}. The computed anomalies \( \mathbf{X}' \), still maintaining the [channel, latitude, longitude] structure, are then standardized by dividing each by its temporal standard deviation \( \sigma \) computed over the same dimensions. This scaling transforms the data into a form where the variance is normalized across the dataset: $ \mathbf{Z} = {\mathbf{X}'}/{\sigma} $
Here, \( \mathbf{Z} \) represents the standardized value, which aligns with the principles of z-score standardization.

The $\mathbf{X}_{\text{clim}}$ and  $\sigma$ values are calculated based on the period from 1980 to 2014 using historical simulations and are specific to each model because each model has different background climate largely due to different model physics and numerical schemes. These postprocessing steps ensure that the dataset is not only cleansed of inherent seasonal biases but also standardized in a manner conducive to extracting meaningful patterns through ML techniques. We acknowledge that alternative normalization/scaling schemes may offer advantages over z-score standardization, and encourage users to explore these options.

\subsection{Dataset Split}
The \texttt{ClimDetect} dataset, encompassing a total of 1,173,913 samples, is carefully divided into training, validation, and testing subsets for effective model training, parameter tuning, and performance evaluation. Specifically, 76.5\% of the samples (897,681 samples) are allocated to the training dataset, 9.9\% (116,727 samples) to the validation dataset, and the remaining 13.6\% (159,505 samples) to the testing dataset. We also provide a "mini" version of the dataset for the purpose of prototyping with 14,986 / 4,244 / 4,244 samples in train / validation / test splits.

When distributing the climate models across these subsets, a key consideration is the "climate sensitivity" of each model \cite{zelinka2020causes, meehl2020context}. Climate sensitivity refers to a model's responsiveness to climate forcings, such as greenhouse gases, which can cause variations in the projected warming. Models vary in their climate sensitivity; some predict higher temperatures under the same forcing scenarios (more sensitive), while others forecast less warming (less sensitive). To ensure a comprehensive and balanced evaluation, we have deliberately selected models across the entire spectrum of climate sensitivity for each dataset split. 

\subsection{Dataset Access}
The \texttt{ClimDetect} dataset is publicly hosted on Hugging Face at \url{https://huggingface.co/datasets/ClimDetect/ClimDetect}.
It is structured using the Hugging Face Datasets library, ensuring full integration within the Hugging Face ecosystem. This integration facilitates seamless interoperability with popular machine learning frameworks (e.g., PyTorch, TensorFlow, and JAX). Moreover, by leveraging the standardized formats and APIs provided by the Hugging Face Datasets library, \texttt{ClimDetect} provides a user-friendly dataset that emphasizes both efficiency and reproducibility.

\section{Framework for Climate Change Detection}
\label{sec:detection_method}
Our climate change detection framework in \texttt{ClimDetect} builds upon the approach of Sippel et al. \citep{Sippel.2020} but distinguishes itself by employing modern AI architectures rather than traditional ridge regression models. For a visual overview, see Figure \ref{fig:cartoon1}.

\textbf{Step 1: Climate Change Detection Model Training}.
The detection task presented by \texttt{ClimDetect} is essentially a regression problem with mapping from a multivariate input tensors to a scalar, where the input is $\mathbf{X} \in \mathbb{R}^{64 \times 128 \times 3}$
and the output is $y \in \mathbb{R}$. Detection models in most prior studies focus on extracting `fingerprints'---characteristic spatial patterns anticipated to emerge due to external forcings such as greenhouse gas emissions. With the application of nonlinear machine learning models, these `fingerprints' are reinterpreted as complex nonlinear functions. These models are trained to discern anthropogenic climate signals from the natural variability present in daily climate data. Specifically, these functions ($F_{\theta}$) are trained on the CMIP6 dataset to map input \textit{daily} climate fields ($\mathbf{X}$) to a \textit{annual} scalar target variable ($y$), a key climate change indicator, establishing a model for climate change signal detection, i.e., $y = F_{\theta}(\mathbf{X})$. 

\textbf{Step 2: Hypothesis Testing}.
The null hypothesis posits that the predicted test statistic falls within the range expected under natural variability. We estimate the distribution of natural variability, $P(y_{\text{hist}}) = P(F_{\theta}(\mathbf{X}_{\text{hist}}))$,\footnote{Subscripts “hist” and “obs” indicate historical and observational data, respectively.} by predicting test statistics from the historical (i.e., the ``pre-warming'' period prior to the onset of significant anthropogenic warming) CMIP6 dataset for the period 1850--1949. Then, we apply the trained model to reanalysis datasets to obtain observed test statistics $y_{\text{obs}} = F_{\theta}(\mathbf{X}_{\text{obs}})$. Finally, we test the null hypothesis by assessing if $y_{\text{obs}}$ is distinguishable from the estimated natural variability, e.g.,  2.5th--97.5th percentile range of $P(y_{\text{hist}})$.

\textbf{Year of Emergence}. We quantify hypothesis testing outcomes with the Year of Emergence (YoE), an important metric for climate projections and policy planning. YoE is defined as the first year when climate change signals statistically surpass daily natural variability (Figures \ref{fig:si_yoe_vit_tas_mr} and \ref{fig:cmip}). An earlier YoE indicates a more sensitive detection model, implying better performance in extracting climate change signals. For robust detection, we establish an ad-hoc threshold for the emergence fraction (EF; defined as the ratio of days on which climate change is detected to the total days in a year) at 97.5\%, equivalent to 356 days.

\section{Benchmark}
\subsection{Experiments}
\label{sec:benchmark}
The benchmark experiments designed for the \texttt{ClimDetect} dataset (Table \ref{tab:exp_res}) aim to comprehensively evaluate the effectiveness of using different combinations of climate variables to predict AGMT. The primary experiment, named "tas-huss-pr," utilizes all three \texttt{ClimDetect} variables—surface temperature, surface humidity, and total precipitation rate—as inputs to estimate AGMT. This experiment serves as the foundation for understanding the combined predictive power of these variables.

In addition to the "tas-huss-pr" experiment, we conducted a series of supplementary experiments to explore the predictive utility of individual variables, reflecting common approaches in prior studies that used a single climate variable as input. These experiments are categorized under the single-variable setup, where each experiment uses only one of the three available variables. Specifically, we have the "tas{\_}only" experiment using only surface temperature, the "huss{\_}only" experiment focusing solely on surface humidity, and the "pr{\_}only" experiment that considers only the total precipitation rate. Each of these experiments provides insights into the individual contributions of the variables to the accuracy of AGMT predictions.

Furthermore, we included two "mean-removed" ("mr") experiments, which are modifications of the "tas-huss-pr" and "tas{\_}only" setups. In these experiments, the spatial mean of each climate field snapshot is removed before conducting the analysis. The rationale behind these "mean-removed" experiments stems from the work of Santer et al. \citep{Santer.2018} and Sippel et al. \citep{Sippel.2020}, which suggest that removing the global mean changes can reveal more about the spatial patterns that contribute to climate signal detection. This approach is predicated on the idea that focusing on spatial anomalies, rather than spatially-averaged residual values, can enhance the detection of climate change signals based on the similarity of spatial patterns alone, thereby increasing confidence in the detection outcomes.

\begin{table}
  \label{experiments-table}
    \caption{Summary of Benchmark Experiments for \texttt{ClimDetect} Dataset. %
    }
  \centering
  \begin{tabular}{llll}
    \toprule
    \makecell[l]{Experiment \\ Name}           & \makecell[l]{Input \\ Variable}                      & \makecell[l]{Target \\ Variable} & \makecell[l]{Mean \\ Removed} \\
    \midrule
    tas-huss-pr                      & tas, huss, pr                        & AGMT            & No           \\
    tas\_only                  & tas                                  & AGMT            & No           \\
    huss\_only                 & huss                                 & AGMT            & No           \\
    pr\_only                   & pr                                   & AGMT            & No           \\
    tass-huss-pr\_mr         & tas, huss, pr                        & AGMT            & Yes          \\
    tas\_mr     & tas                                  & AGMT            & Yes          \\
    \bottomrule
  \end{tabular}
  \label{tab:exp_res}
\end{table}

\subsection{Baseline Models and Training Details}
\label{baseline-training-details}
Using the \texttt{ClimDetect} dataset, we apply a collection of vision transformer (ViT) based models, which has not been tested for climate change detection problems. To do this, we add a regression head to a ViT and jointly train the regression head and model on the train / validation splits of the \texttt{ClimDetect} dataset. We test four popular ViTs: ViT-b/16 \citep{dosovitskiy2020image}, CLIP \citep{radford2021learning}, MAE \citep{he2021masked}, and DINOv2 \citep{oquab2024dinov2}. In addition, we include ResNet-50 \citep{he2016deep} as a convolutional neural network (CNN) baseline, which is another widely used model for computer vision tasks and was also evaluated in a previous climate D\&A study \citep{Ham.2023}.

These ViT models are widely used in the computer vision and multimodal literatures \citep{awais2023foundational}.
Adjustments specific to our climate data involved training with a uniform batch size of 64 and a learning rate of 5e-4 (with warm-up and decay), optimized through a grid search initially conducted on Google’s ViT-b/16 for the "tas-huss-pr" setup. 
We trained for 10 epochs using the AdamW optimizer updating all parameters of the model. Total training took 4.75 hours on average per model using eight A6000 Nvidia GPUs on an internal Linux Slurm cluster. 
We provide additional details on the training in the supplementary materials.

To provide a comprehensive evaluation, traditional models used in climate science were also included: a ridge regression model and a multilayer perceptron (MLP). The ridge regression was tuned with a large alpha value ($10^6$), while the MLP featured five hidden layers with 100 units each, a learning rate of 5e-5 with cyclic learning rate adjustments, and L2 regularization with $\alpha$=0.01. These models served as benchmarks to assess the state-of-the-art capabilities of ViT models against conventional methods in the context of climate change detection tasks.

\subsection{Results}
In assessing the performance of our baseline models on the withheld test split, we used RMSE as the primary evaluation metric—one of the most widely used metrics in climate science. RMSE has proven to be a reliable proxy for detection sensitivity. For instance, improvements in RMSE correlate with enhanced sensitivity by tightening the distribution of both the reference and test periods \citep{Sippel.2020}.

Our RMSE analysis across six experiments with seven baseline models reveals a competitive landscape with relatively similar performance levels (Table \ref{tab:RMSE_test}). In most experiments involving multiple variables (e.g., "tas-huss-pr" and "tas-huss-pr\_mr"), at least one of the four ViT baselines outperforms the non-ViT models (MLP, CNN, and ridge regression), though the specific ViT model showing superior performance varies across experiments. With the notable exception of the "tas\_mr" experiment—where both MLP and ridge regression outperformed the ViT and CNN models—these findings suggest that ViTs are generally better suited for modeling the complex, high-order interactions among multiple climate variables. In particular, the strong performance of a ViT model in the most challenging experimental setup ("tas-huss-pr\_mr"), where the model must capture multi-variable interactions without the mean signal, underscores their potential for developing even more sensitive climate change detection models. One possible explanation for the observed performance divergence is that the ViT hyperparameters were tuned on the ViT-b/16 model with the "tas-huss-pr" configuration, which may favor that specific setup over others.

In contrast, in the pr\_only experiment---where models relied solely on precipitation data---all models, particularly ridge regression, struggled, likely due to the sparse and indirect relationship between precipitation and other climate state variables (e.g., temperature and humidity).

Overall, these findings underscore the potential of ViTs in future climate detection and attribution studies, especially in scenarios that involve multiple variables and complex data configurations. Nonetheless, the optimal model choice should consider the specific requirements and characteristics of each experiment.

\begin{table}
\centering
\caption{Root Mean Square Error (RMSE) across different models and experiments, calculated over the ClimDetect test set that spans 150 years (1950-2100) [Unit: $^\circ$C]. RMSE values are underlined if their 95\% confidence interval, determined by resampling the test set with replacement 10,000 times, overlap with that of the best-performing model. "t-h-p" abbreviates the tas-huss-pr experiment.}
\resizebox{\columnwidth}{!}
{
\begin{tabular}{lcccccc}                                                \toprule
& t-h-p          & tas            & pr             & huss           & \makecell{t-h-p\\(mr)}      & \makecell{tas\\(mr)}              \\ \midrule
CLIP                 & \textbf{0.1411}      & \textbf{0.1482}               & 0.8935               & 0.1801               & 0.1690               & \underline{0.2410}               \\
DINOv2               & 0.1439               & 0.1645               & 0.7995               & 0.1942               & 0.1731               & 0.2552               \\
MAE                  & 0.1430               & \underline{0.1484}        & 0.6451               & \textbf{0.1571}               & \textbf{0.1672}               & 0.2531               \\
ViT-b/16                  & 0.1425          & 0.1610               & 0.7132               & 0.1604               & 0.1763               & 0.2562               \\
ResNet-50            & 0.1471               & 0.1687               & \textbf{0.6137}               & 0.1661               & 0.1835               & 0.2693               \\
MLP                  & 0.1488               & 0.1557               & 0.7502               & 0.1804               & 0.2192               & \underline{0.2409}               \\
ridge                & 0.1508               & 0.1542               & 0.9708               & 0.2304               & 0.2156               & \textbf{0.2404}               \\ \bottomrule
\end{tabular}
}
\label{tab:RMSE_test}
\end{table}

\begin{table}
\centering
\caption{RMSE calculated over the most recent 45 years (1980-2024) of ERA-5 reanalysis data [Unit: $^\circ$C]. (For details on highlighted RMSE values and the abbreviation "t-h-p", see Table \ref{tab:RMSE_test} caption.) Corresponding RMSE tables for JRA-3Q and MERRA-2 are presented in Appendix Tables \ref{tab:RMSE_JRA3Q} and \ref{tab:RMSE_MERRA-2}.}

\resizebox{\columnwidth}{!}
{
\begin{tabular}{lcccccc}
\toprule
& t-h-p & tas & pr & huss & \makecell{t-h-p\\(mr)} & \makecell{tas\\(mr)} \\
\midrule
CLIP        & 0.1064 & 0.1291 & \textbf{0.5269} & 0.1925 & 0.1785 & 0.1873 \\
DINOv2      & 0.1119 & 0.1387 & 0.5890 & 0.1924 & 0.1595 & 0.1797 \\
MAE         & \textbf{0.0921} & 0.1041 & 0.7656 & \textbf{0.1321} & \textbf{0.1308} & 0.1508 \\
ViT-b/16    & 0.1031 & \textbf{0.0839} & 1.0125 & 0.1695 & \underline{0.1331} & \textbf{0.1433} \\
ResNet-50   & 0.0959 & 0.0882 & 0.6458 & 0.1801 & 0.1596 & 0.1821 \\
MLP         & 0.0995 & 0.1077 & 0.6141 & 0.1631 & 0.1698 & 0.1720 \\
ridge       & \underline{0.0943} & 0.1001 & \underline{0.5372} & 0.1894 & 0.1496 & 0.1796 \\
\bottomrule
\end{tabular}
}
\label{tab:RMSE_ERA5}
\end{table}

\subsection{Detecting Climate Change Signal from Real-World Observation Data}
We used three reanalysis datasets (ERA5, JRA-3Q, and MERRA-2) to test our baseline models in detecting climate change signals from real-world observation data. We begin by analyzing RMSE and then examine the year of emergence (YoE) in the following section.

Despite subtle differences, the RMSE on ERA5 broadly aligns with that on the \texttt{ClimDetect} test set---which comprises CMIP6 climate model simulation data---indicating that at least one of the ViT baselines performs better in most experiments, except in the “pr” experiment (Table \ref{tab:RMSE_ERA5}). While MAE and ViT-b/16 consistently show low RMSE for most variables except “pr,” CLIP and DINOv2 do not uniformly outperform simpler models like MLP and Ridge Regression, particularly in configurations such as “tas-huss-pr,” “huss,” and “tas\_mr.”

The RMSE in JRA-3Q and MERRA2 (shown in Tables \ref{tab:RMSE_JRA3Q} and \ref{tab:RMSE_MERRA-2}) echoes similar findings, with ViTs generally outperforming the non-ViT models. However, the specific ViT model that performs best varies across the three reanalysis datasets. Apart from differences in hyperparameter optimization, these discrepancies likely arise from variations in the assimilation models and observational inputs used in these systems.

Additionally, RMSE values are generally lower on the reanalysis data than on the CMIP6 data, likely due to differences in the evaluation periods rather than in model generalization. For example, uncertainties in the CMIP6 output increase over the projection period (Figure \ref{fig:cmip}).

\begin{figure} %
    \centering 
    \includegraphics[width=\columnwidth,trim=0cm .3cm 0cm 0cm, clip]{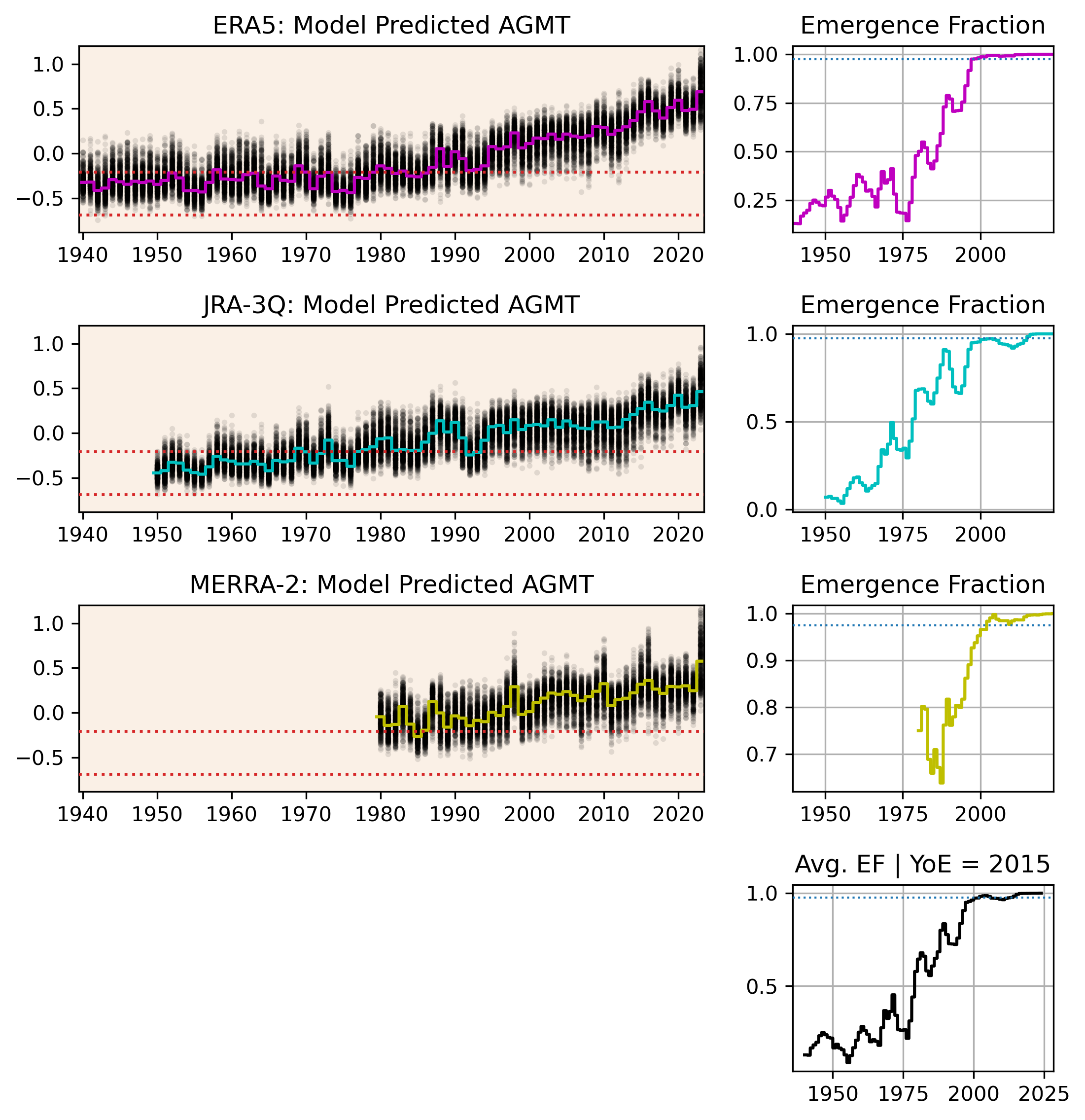}
    \caption{Detection model: \textbf{ViT-b/16}; Experiment: "tas\_mr". (Left) Model-predicted test statistic, AGMT, from three different reanalysis datasets, displayed as 365 black dots per year with their mean represented by the colored line. The red lines indicate the 2.5th to 97.5th percentile range of natural variability for the test statistic, which was estimated from the 1850-1949 CMIP6 model simulation output (the test split). (Right) Emergence fraction (EF) per year, defined as the fraction of days where predicted AGMT exceeds the upper bound (the 97.5th percentile of natural variability) within one year. Centered 5-year window moving averaging is applied to EF time series. (Bottom Right) The black line represents the average of the three colored lines shown in the upper panels. The Year of Emergence (YoE) is calculated from this average, defined as the first year where the averaged EF surpasses the 97.5\% threshold (blue line), corresponding to 356 days.} 
    \label{fig:si_yoe_vit_tas_mr}
\end{figure}

\subsection{Year of Emergence}
\label{sec:si_YOE}
Next, we examine the Year of Emergence (YoE), a task-specific metric for climate change detection. Thus far, we have evaluated baseline model performance using RMSE (Step 1 in Section \ref{sec:detection_method}); however, we have not yet assessed whether a given daily input snapshot contains a detectable climate change signal using the observation dataset (Step 2 in Section \ref{sec:detection_method}). Here, we test whether the observational daily snapshot exhibits a climate change signal, and we determine the year in which this signal robustly emerges (See Figure \ref{fig:si_yoe_vit_tas_mr} for details).

\begin{figure}
    \centering
        \includegraphics[width=0.75\columnwidth,trim=.2cm .25cm 0 .7cm, clip]{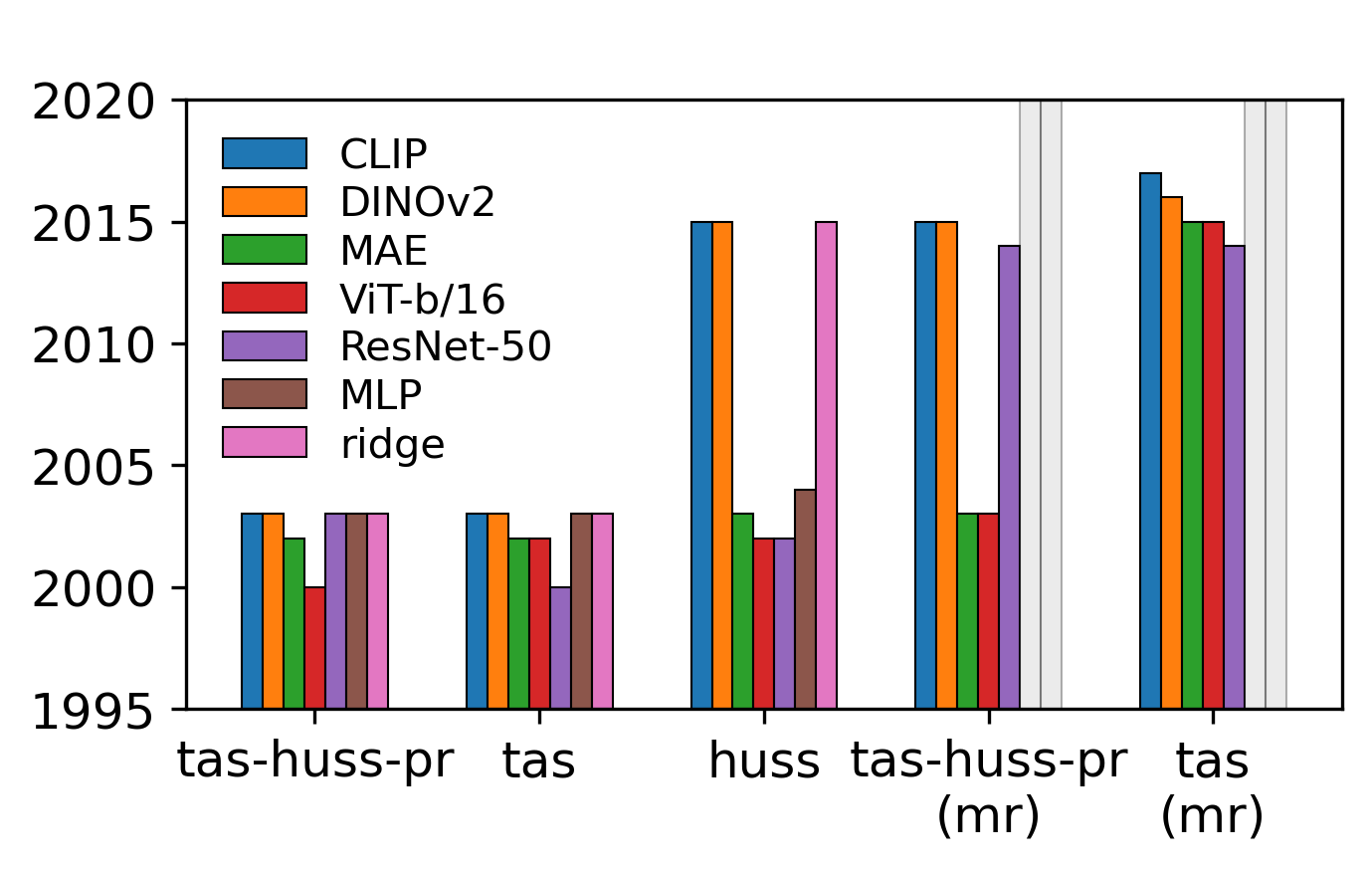}
    \caption{Year of emergence (YoE), defined as the first year when at least 97.5\% of daily climate fields show a distinguishable climate change signal from natural variability. Grey bars indicate instances where a model failed to capture YoE within the reanalysis period of 1980-2024. "pr" is omitted since no detection model can capture YoE. 
    } 
    \label{fig:YOE}
\end{figure}

In contrast to RMSE, YoE distinctly highlights the effectiveness of sophisticated models such as ViTs and CNNs (Figure \ref{fig:YOE}). Across all experiments, MAE, ViT-b/16, and ResNet-50 consistently exhibit the earliest YoE, indicating their higher detection sensitivity. Conversely, ridge regression and MLP perform comparably to less effective ViTs (such as CLIP and DINOv2) when evaluated using RMSE (Table \ref{tab:RMSE_ERA5}), but fail to detect an emergence in the mean-removed experiments ("tas\_mr" and "tas-huss-pr\_mr") at the 97.5\% emergence fraction (EF) threshold. This finding is consistent across various EF thresholds (Figure \ref{fig:si_yoe_th}), providing further evidence of the potential of ViTs to improve current climate change detection models.

\begin{figure*}
    \centering
    \includegraphics[width=.85\textwidth,trim=0 1.05cm 0 .7cm, clip]{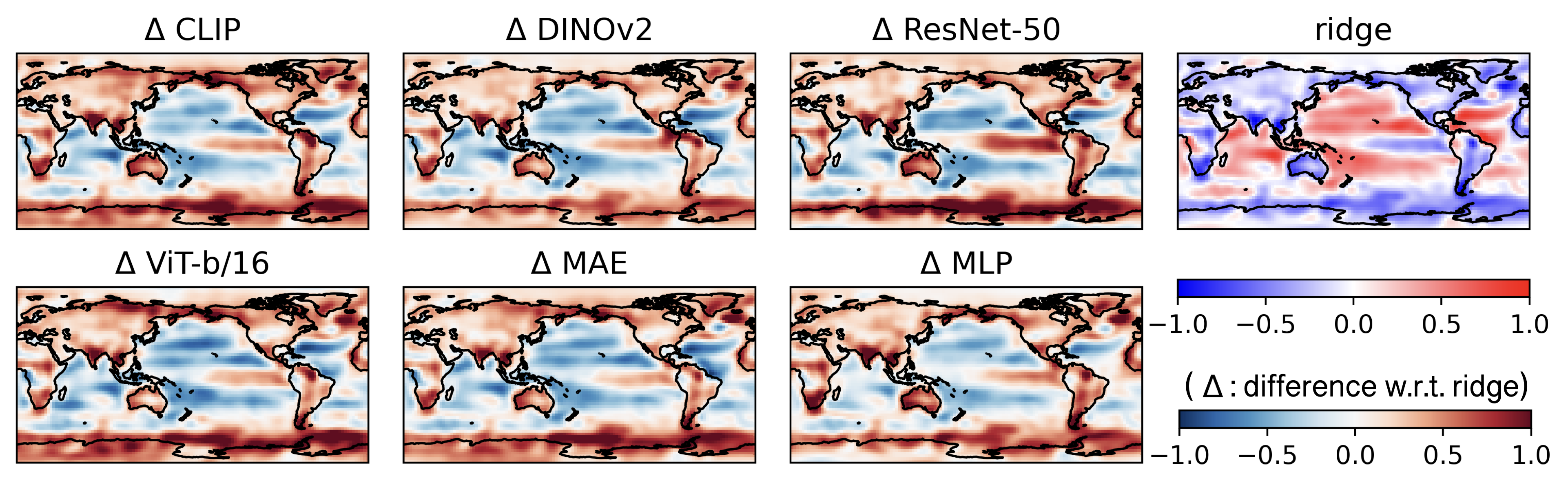}
    \caption{Visualization of Integrated Gradients (IG) times Input for the "tas-huss-pr\_mr" experiment, highlighting regions influencing the prediction of AGMT. Appendix \ref{App:IG} includes IG$\times$Input visualizations for other experiments.}
    \label{fig:IG}
\end{figure*}

\subsection{Physical Interpretation}
\label{Sec:interpretation}
Physical interpretability remains crucial for establishing data-driven models as a valuable tool in climate science. We present preliminary model interpretations using Integrated Gradients (IG) \cite{sundararajan2017axiomatic}. To facilitate this analysis, we collected approximately 26k samples from the \texttt{ClimDetect} test set for which the target AGMT falls within the [1.5, 2.5) bin, a range representative of significant climate change (Figure \ref{fig:cmip}). For these samples, IG$\times$Input values were computed, averaged, smoothed using a Gaussian filter, and normalized by the maximum IG$\times$Input value. 

In addition, we visualized ridge regression coefficients—which indicate the linear sensitivity of local climate variables to the target—as a first-order baseline for interpretability. For the "tas-huss-pr\_mr" experiment, where a pronounced performance gap between simple and advanced models is observed, Figure \ref{fig:IG} displays the differences ($\Delta$) with respect to the ridge regression model for all models except ridge.

This figure reveals distinct differences between nonlinear ML models and ridge regression. Unlike ridge, ViTs (along with CNN and MLP) exhibit a diminished focus on land-sea contrasts and a stronger positive dependence on the Antarctic Ocean. These consistent patterns across different architectures suggest that ViTs may better capture underlying physical processes. Overall, these findings highlight the nuanced capabilities of ViTs—and the machine learning approach in general—in advancing climate detection and attribution.

\section{Limitations and Future Work}
\label{limitations}

\textbf{Data Selection}. We acknowledge that our dataset does not encompass the complete range of CMIP6 model outputs. Some models were omitted due to server errors and availability issues at the time of data preparation. Despite these omissions, we believe their impact on our overall findings is minimal. We also plan to regularly update our dataset as significant new models become available in the CMIP6 archives.

\textbf{Baseline Model Hyperparameters}. Due to computational constraints, we tuned the hyperparameters for the ViT models using grid search on the ViT/b-16 model with the "tas-huss-pr" configuration and then applied these hyperparameters uniformly across all ViT models and experimental setups. While this approach was necessary given our limited resources, it likely influenced our benchmark results. Future work should investigate more targeted hyperparameter tuning for each model and experiment.

\textbf{Model Interpretation}. Interpreting complex machine learning models remains challenging. One limitation in our physical interpretation analysis (Section \ref{Sec:interpretation}) is the difficulty of obtaining a definitive ground truth for model explanations. Although ridge regression coefficients provide a linear baseline for understanding localized, pixel-level relationships, they are inherently limited. Future research should focus on developing methodologies that more accurately capture both linear and nonlinear interactions, and on integrating physically grounded knowledge into the evaluation process.

\section{Conclusion}
We introduced the \texttt{ClimDetect} dataset, a standardized benchmark designed to unify previous efforts in climate change detection and attribution by using consistent input/output variables, data, and models from both historical and ScenarioMIP experiments of CMIP6. In addition, \texttt{ClimDetect} includes three state-of-the-art reanalysis products (ERA5, JRA-3Q, and MERRA-2) for testing and validating detection models. The dataset further provides robust benchmark baselines by incorporating four popular vision transformers---applied for the first time to climate change detection---alongside three established baselines (ridge regression, MLP, and CNN). This comprehensive framework supports end-to-end climate change detection and attribution benchmarking, ensuring reproducibility and comparability across different models. We anticipate that the \texttt{ClimDetect} dataset will not only advance the integration of machine learning in climate science but also lay the groundwork for future research and policy-making aimed at effectively addressing global climate challenges. Although we foresee no significant negative impacts given the nature of the dataset, we are committed to ongoing monitoring to ensure its responsible use.

\bibliographystyle{unsrt}
\bibliography{main}

\appendix

\section{Concepts and Terminology from Climate Science}
\label{sec:appendix_concepts}
To motivate the problem of climate D\&A for the broader ML/DL community, it is important to clarify some fundamental concepts and terminologies. This subsection will introduce core climate science concepts that are crucial for interpreting climate data and projections: natural climate variability, CMIP6 climate projections, and the sources of uncertainties in these projections. Figures \ref{fig:cartoon1} and \ref{fig:cmip} shows these concepts applied to representative data from the ClimDetect dataset. Predictions of a target variable representing climate forcing (Annual Global Mean Temperature--AGMT) are made from a model trained on the ClimDetect dataset given inputs of daily variables from the CMIP6 climate model ensemble. Input data span historical and future climate scenarios, illustrating warming trends, while confidence intervals illustrate the range of climate variability. The sensitivity of a specific model for D\&A can be measured by its ability to reduce the variance in the AGMT prediction relative to the variability of background climate (see Figure 1). 

\subsection{Natural Climate Variability (Internal Variability)}
Natural climate variability, also known as internal variability, refers to the inherent fluctuations in climate parameters caused by the internal dynamics of the Earth's climate system. These fluctuations occur across various timescales, from seasonal to multi-decadal, and are independent of external forcing factors like volcanic eruptions or human-induced greenhouse gas emissions. Such variability is driven by complex interactions within the climate system, including the atmosphere, oceans, cryosphere, and land surfaces. For instance, the El Niño-Southern Oscillation (ENSO) represents a significant pattern of natural variability with substantial impacts on global weather and climate on an interannual scale. Decadal oscillations like the Pacific Decadal Oscillation (PDO) and the Atlantic Multidecadal Oscillation (AMO) also exemplify longer-term internal variability that can modulate global and regional climate trends. Understanding these patterns is crucial for distinguishing between changes in climate due to external forcings and those arising from the climate system's inherent dynamics.

\subsection{CMIP6 Climate Projections}
 CMIP6 is a globally coordinated effort involving over 100 climate models from more than 50 modeling groups, making it one of the most comprehensive climate modeling projects to date. With a total data volume exceeding 20 petabytes, CMIP6 plays a crucial role in the Intergovernmental Panel on Climate Change (IPCC) Assessment Reports (AR). These reports are essential for providing policymakers with standardized climate projections and historical simulations that form the backbone of climate change assessments. The IPCC uses data from
CMIP6 to evaluate climate models, compare their outputs, and produce projections for future
climate scenarios, which inform global climate policy and adaptation strategies (Copernicus GMD)(Copernicus BG) (Copernicus).
The historical simulations in CMIP6 are designed to recreate the climate of the past, typically from around 1850 to 2014. These simulations incorporate a wide range of observed data, including greenhouse gas concentrations, volcanic eruptions, solar radiation, and land use changes, helping scientists understand how natural and human activities have influenced climate changes over the past 150 years. ScenarioMIP, a part of CMIP6, focuses on future climate projections from 2015 to 2100. Based on various socio-economic pathways known as Shared Socioeconomic Pathways (SSPs), these simulations consider different future scenarios like SSP2-4.5 (a moderate scenario) and SSP3-7.0 (a high-emission scenario). By providing a range of potential future climates, ScenarioMIP helps policymakers and researchers explore the implications of different climate action strategies (Copernicus GMD) (Copernicus). This dataset, processed and utilized in our study, leverages the robust and detailed outputs from these simulations to support our research objectives.

\subsection{Sources of Uncertainties in Climate Projections}
The projection of future climate conditions involves several sources of uncertainty that need careful consideration: [1] Natural Variability: The inherent variability within the climate system can mask or enhance climate trends on both short and long timescales; [2] Scenario Uncertainty: This arises from the difficulty in predicting future changes in socio-economic conditions, technological developments, and climate policy, all of which affect greenhouse gas emissions and land use changes; and [3] Model Uncertainty: Different climate models may represent physical processes differently or have different sensitivities to greenhouse gas concentrations, resulting in varied predictions under identical scenarios.

\section{Training details}
\label{Sec:si_training_details}
\textbf{Vision Transformers}. We adopted four Vision Transformer (ViT) models---ViT-b/16, CLIP, MAE, and DINOv2---as described in the ClimDetect baseline models, adhering to their specified configurations and training settings. These models were sourced from Hugging Face (google/vit-base-patch16-224, openai/clip-vit-large-patch14-336, facebook/vit-mae-base, and facebook/dinov2-large, respectively). Each model was trained with a regression head  (that is, \texttt{num\_labels=1}) using a batch size of 512. The learning rate was set at 5e-4, with a warm-up period during the first half of an epoch followed by a fixed linear decay at 5\% for the remainder of the training. The models were trained over 10 epochs using the AdamW optimizer, with all parameters being updated during training. We used the best checkpoints based on the lowest validation loss.

\textbf{CNN}. We chose the ResNet-50 architecutre for our CNN model. ResNet-50 was trained from a Hugging Face (microsoft/resnet-50) with a regression head. The effective batch size was 64. The learning rate was set at 1e-4 with a warm-up period over the first epoch followed by a 5\% linear decay for the remaining epochs. The training was conducted over 10 epochs, and then the best checkpoints were selected based on validation loss.

\textbf{MLP and Ridge Regression}. A ridge regression model was fit with $\alpha=10^6$, and a multilayer perceptron (MLP) featured five hidden layers, each with 100 units. The MLP's learning rate was set at 5e-5 with cyclic adjustments and included L2 regularization set at $\alpha=0.01$.

\begin{figure} %
        \centering
        \includegraphics[width=.7\columnwidth, trim=0cm 0.3cm 0cm 0.1cm, clip]{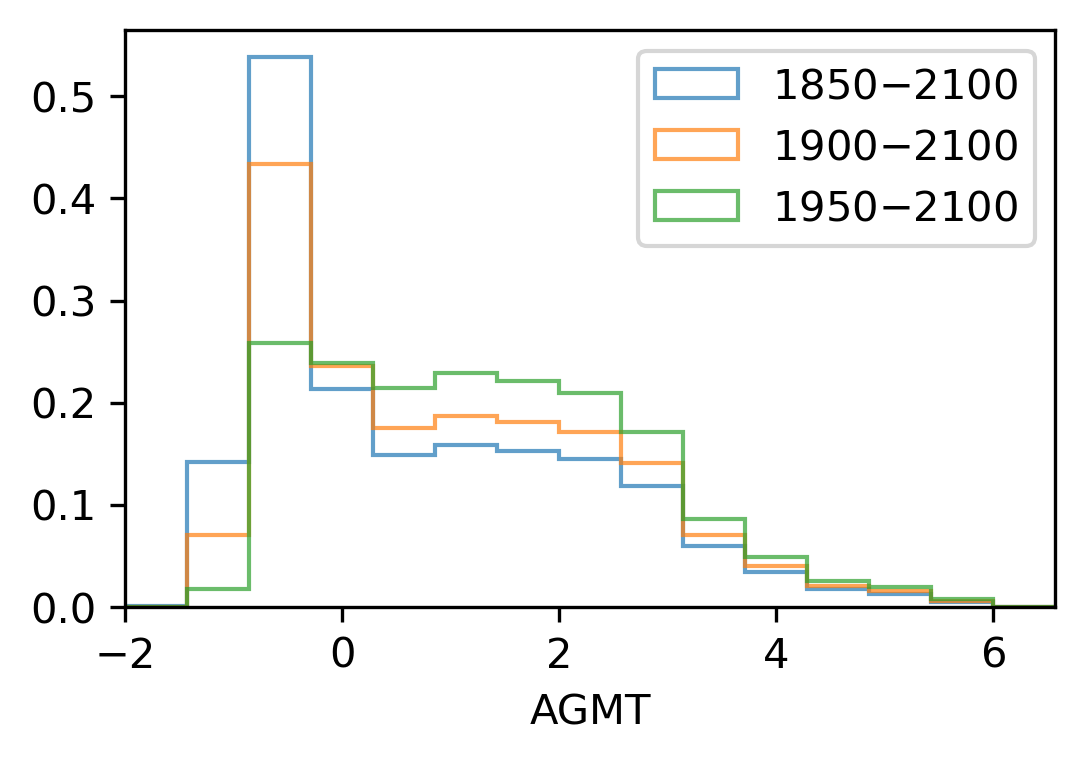}
        \caption{Probability density of AGMT in the training split across three different time periods: (blue) 1850–2100, (orange) 1900–2100 (orange), and (green) 1950–2100.}
        \label{fig:SI_agmt_hist}
\end{figure}

\begin{figure} %
        \centering
        \includegraphics[width=1.\columnwidth, trim=1.6cm 0cm 2.2cm 1cm, clip]{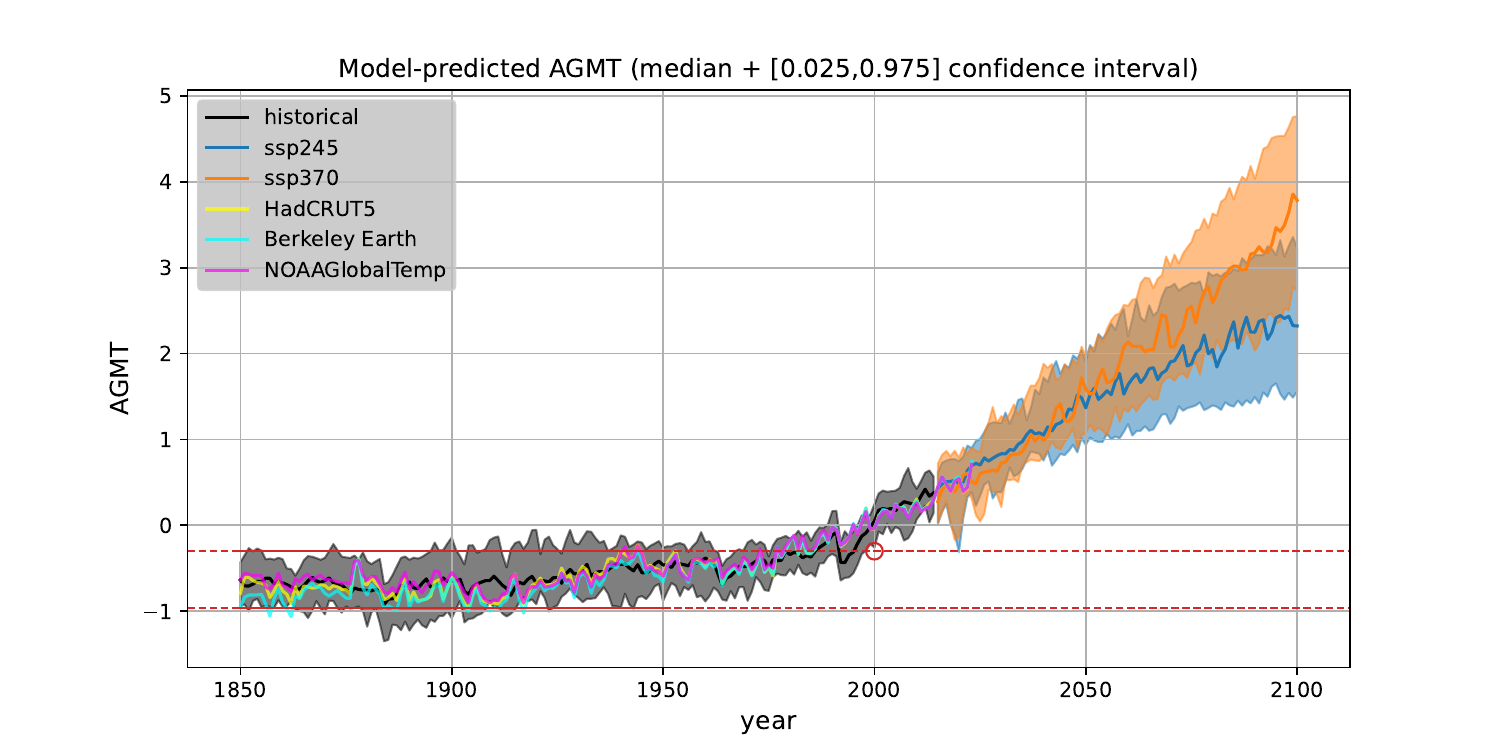}
        \caption{Annual global mean temperature over time from observations (HadCRUT5\cite{hadcrut5}, Berkeley Earth\cite{berkley-earth}, NOAAGlobalTemp\cite{noaaglobaltemp}) and from model (ViT-b/16) predictions made from daily weather variables from CMIP6 models in the \texttt{ClimDetect} test set over historical and projected (SSP2-4.5, SSP3-7.0) climate. The values are relative to the 1980-2014 average. Shaded regions represent the [2.5\%, 97.5\%] confidence intervals from the model prediction. The mean of the CI over the historical period 1850-1950 (solid red line and extended dashed red line) represents the range of the baseline pre-industrial climate and the CI range represents background climate variability. The red circle illustrates the time period during which the warming signal emerges from the range of background climate variability -- even from a daily weather snapshot \cite[e.g.,][]{Sippel.2020, Sippel.2021, Ham.2023}. Models which can reduce the variance in the AGMT target prediction relative to the internal climate variability will be more sensitive to this emergence and hence be more accurate in D\&A. Note that this figure is constructed purely on \texttt{ClimDetect}'s CMIP6 climate model simulation dataset.}
        \label{fig:cmip}
\end{figure}

\textbf{Training Dataset}. We used the training and validation splits for model training. To achieve a balanced distribution of the target variable (AGMT), we restricted our training data to the period 1950–2100 (Figure \ref{fig:SI_agmt_hist}).

\FloatBarrier
\section{RMSE calculated on JRA-3Q and MERRA-2}
\label{sec:si_rmse_Table}

\begin{table} %
\centering
\caption{Similar to Table \ref{tab:RMSE_ERA5} in the main text, but with RMSE calculated over the 1980-2024 period using JRA-3Q data.}
\resizebox{\columnwidth}{!}
{
\begin{tabular}{lcccccc}
\toprule
            & t-h-p & tas & pr & huss & \makecell{t-h-p\\(mr)} & \makecell{tas\\(mr)} \\
\midrule
CLIP        & 0.1166 & 0.1261 & \textbf{0.4518} & 0.1778 & 0.1996 & 0.2283 \\
DINOv2      & 0.1287 & 0.1239 & 0.4724 & 0.1744 & 0.1745 & 0.2438 \\
MAE         & \underline{0.1082} & \underline{0.1020} & \underline{0.4601} & \textbf{0.1301} & \textbf{0.1575} & 0.2072 \\
ViT-b/16    & 0.1310 & \textbf{0.0994} & 0.5293 & 0.1615 & 0.1666 & 0.1870 \\
ResNet-50   & 0.1214 & 0.1230 & 0.5778 & 0.1543 & 0.1947 & \textbf{0.1692} \\
MLP         & 0.1151 & 0.1264 & 0.7135 & 0.1525 & 0.2185 & 0.2270 \\
ridge       & \textbf{0.1074} & 0.1159 & 0.5303 & 0.1671 & 0.1766 & 0.2213 \\
\bottomrule
\end{tabular}
}
\label{tab:RMSE_JRA3Q}
\end{table}

\begin{table} %
\centering
\caption{Similar to Table \ref{tab:RMSE_ERA5} in the main text, but with RMSE calculated over the 1980-2024 period using MERRA-2 data.}

\resizebox{\columnwidth}{!}
{
\begin{tabular}{lcccccc}
\toprule
            & t-h-p & tas & pr & huss & \makecell{t-h-p\\(mr)} & \makecell{tas\\(mr)} \\
\midrule
CLIP        & 0.1284 & 0.1596 & \textbf{0.5281} & 0.2053 & 0.1668 & 0.2407 \\
DINOv2      & 0.1328 & 0.1778 & 0.5784 & 0.2060 & 0.1734 & 0.2461 \\
MAE         & \underline{0.1137} & 0.1372 & 0.6040 & 0.1780 & 0.1463 & 0.1920 \\
ViT-b/16    & \textbf{0.1125} & \textbf{0.1165} & 0.8052 & 0.1770 & \textbf{0.1391} & \textbf{0.1817} \\
ResNet-50   & 0.1196 & 0.1221 & \underline{0.5371} & \textbf{0.1433} & 0.1749 & 0.2375 \\
MLP         & 0.1302 & 0.1340 & 0.6790 & 0.1768 & 0.3002 & 0.2783 \\
ridge       & 0.1257 & 0.1260 & 0.6307 & 0.1770 & 0.2644 & 0.2608 \\
\bottomrule
\end{tabular}
}
\label{tab:RMSE_MERRA-2}
\end{table}

\FloatBarrier
\section{Year of Emergence}

\begin{figure} %
    \centering 
    \includegraphics[width=\columnwidth,trim=0cm .3cm 0cm 0cm, clip]{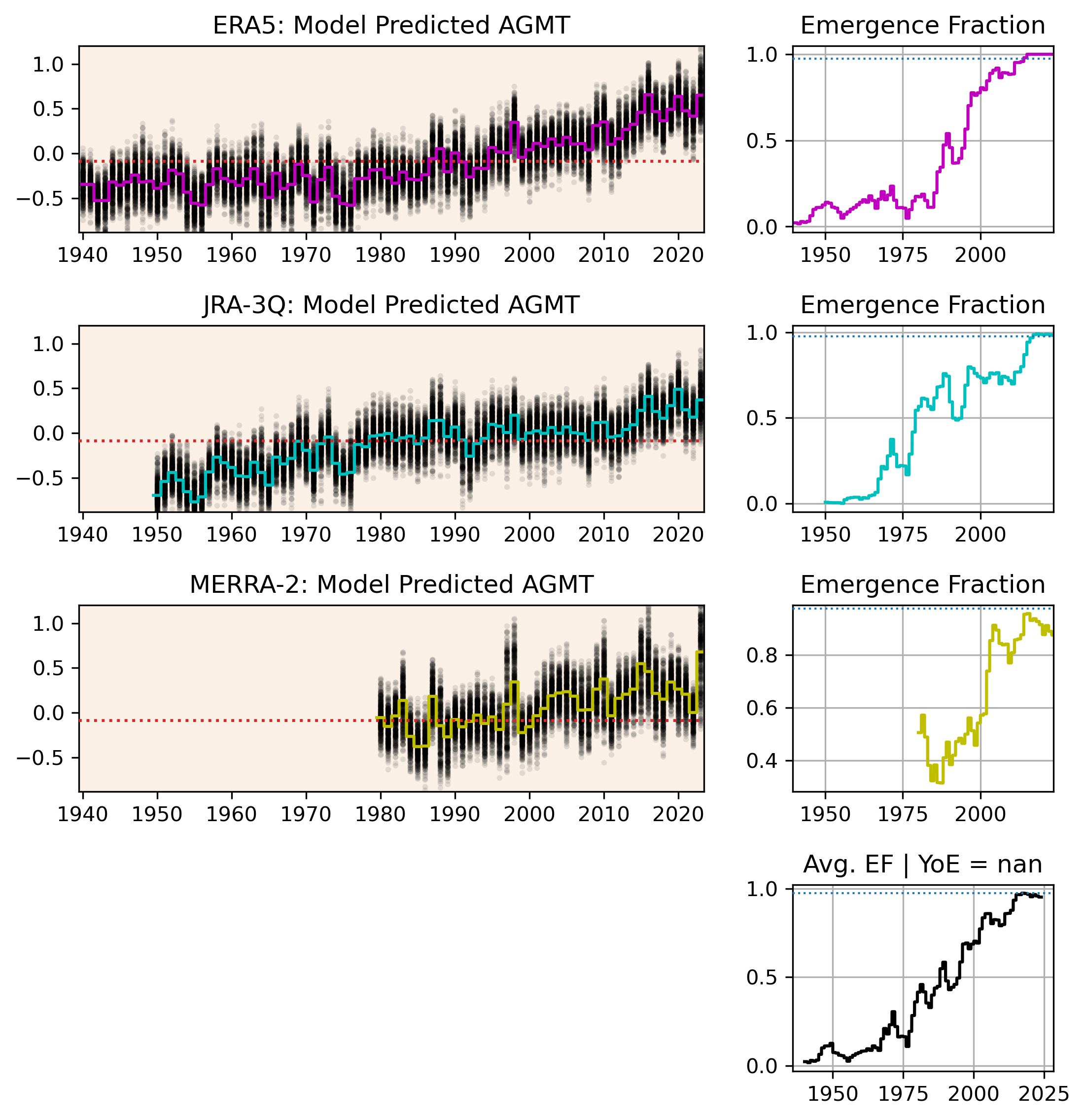}
    \caption{Similar to Figure \ref{fig:si_yoe_vit_tas_mr} but \textbf{Ridge regression} is used as a climate change detection model.} 
    \label{fig:si_yoe_ridge_tas_mr}
\end{figure}

\onecolumn

\begin{figure*} %
    \centering
    \includegraphics[width=0.85\textwidth,trim=0cm 0 0cm 0, clip]{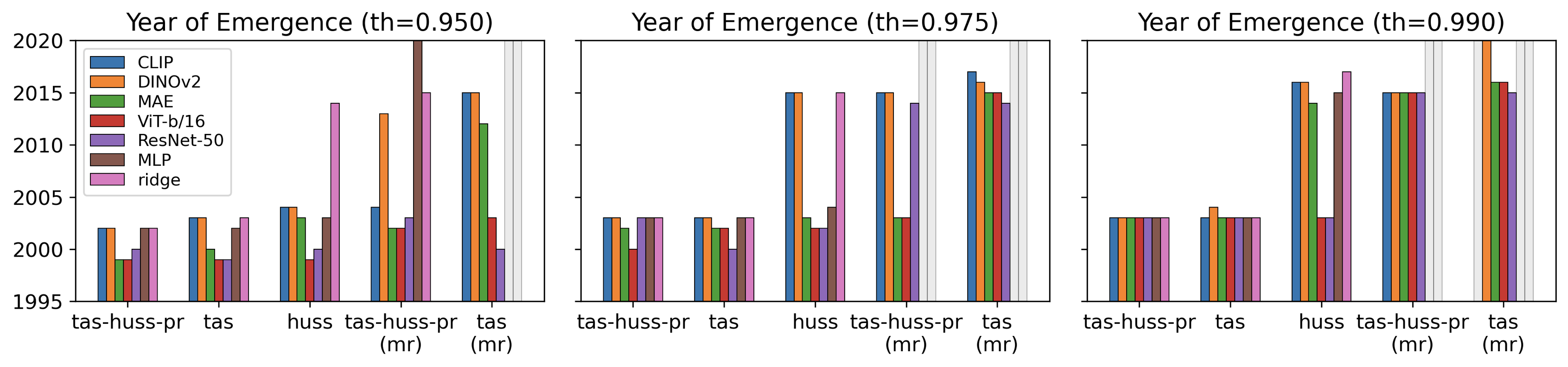}
    \caption{Similar to Figure \ref{fig:YOE}, but with three different emergence fraction threshold: (left) 0.95, (middle) 0.975, and (right) 0.99.}
    \label{fig:si_yoe_th}
\end{figure*}

\FloatBarrier

\section {Integrated Gradients Maps}
\label{App:IG}
\begin{figure*}[h]
    \centering
    \includegraphics[width=.9\linewidth]{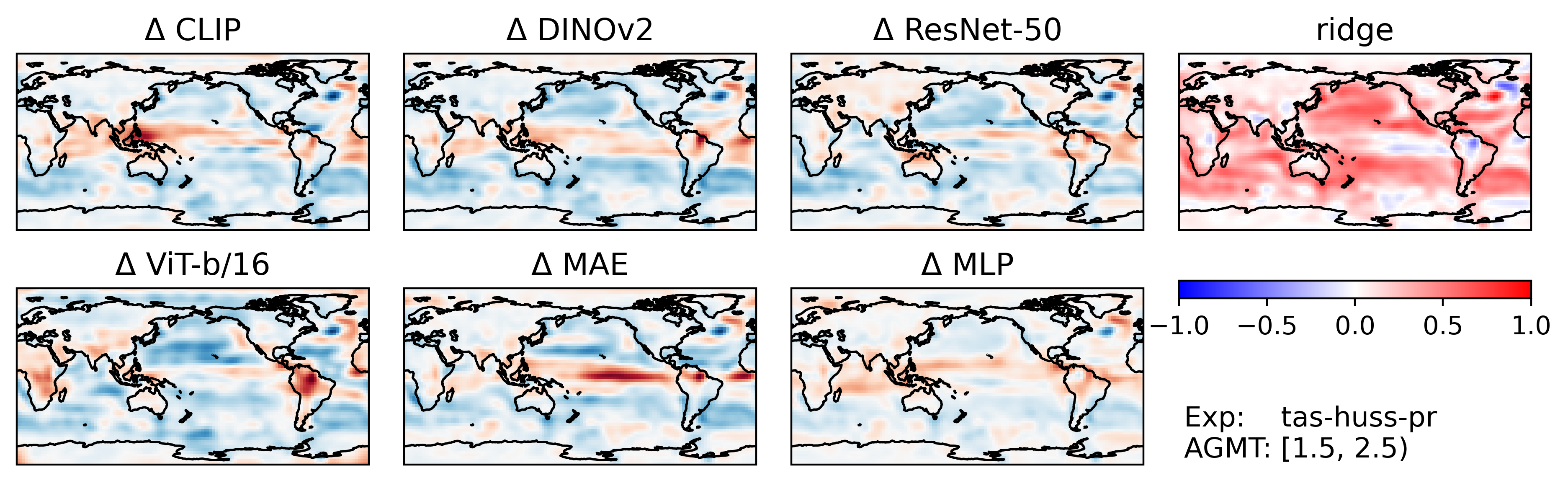}
    \caption{Similar to Figure \ref{fig:IG}, except for the \textbf{tas-huss-pr} experiment}
    \label{fig:si_IG_3var}
\end{figure*}

\begin{figure*}[h]
    \centering
    \includegraphics[width=.9\linewidth]{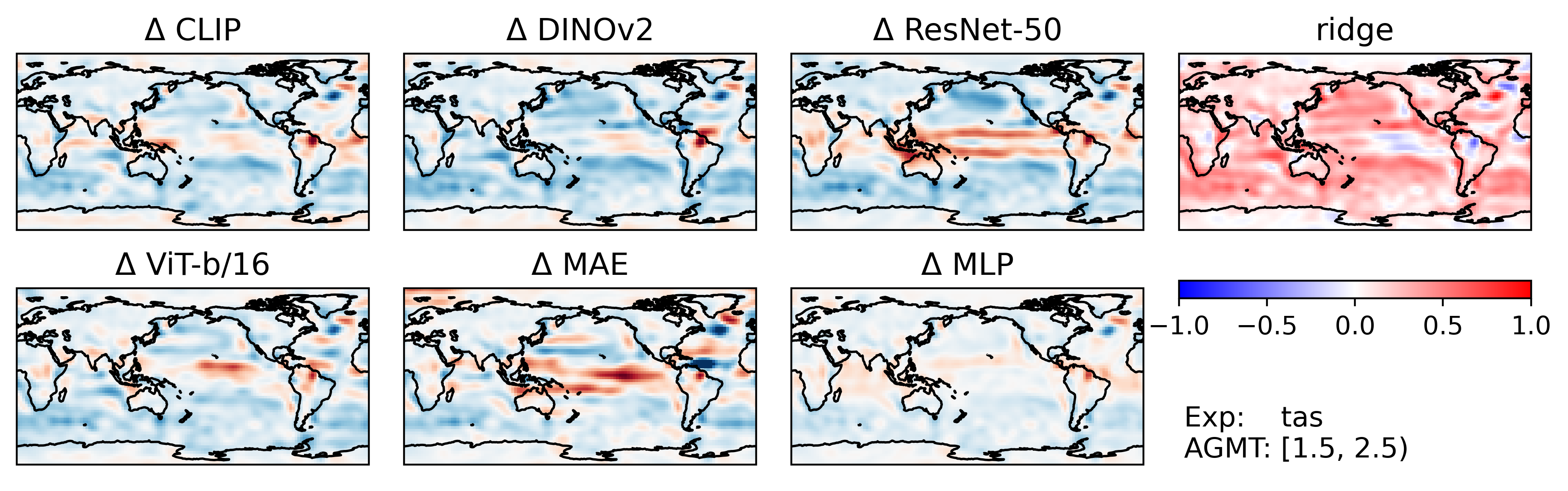}
    \caption{Similar to Figure \ref{fig:IG}, except for the \textbf{tas} experiment}
    \label{fig:si_IG_tas}
\end{figure*}

\begin{figure*}[h]
    \centering
    \includegraphics[width=.9\linewidth]{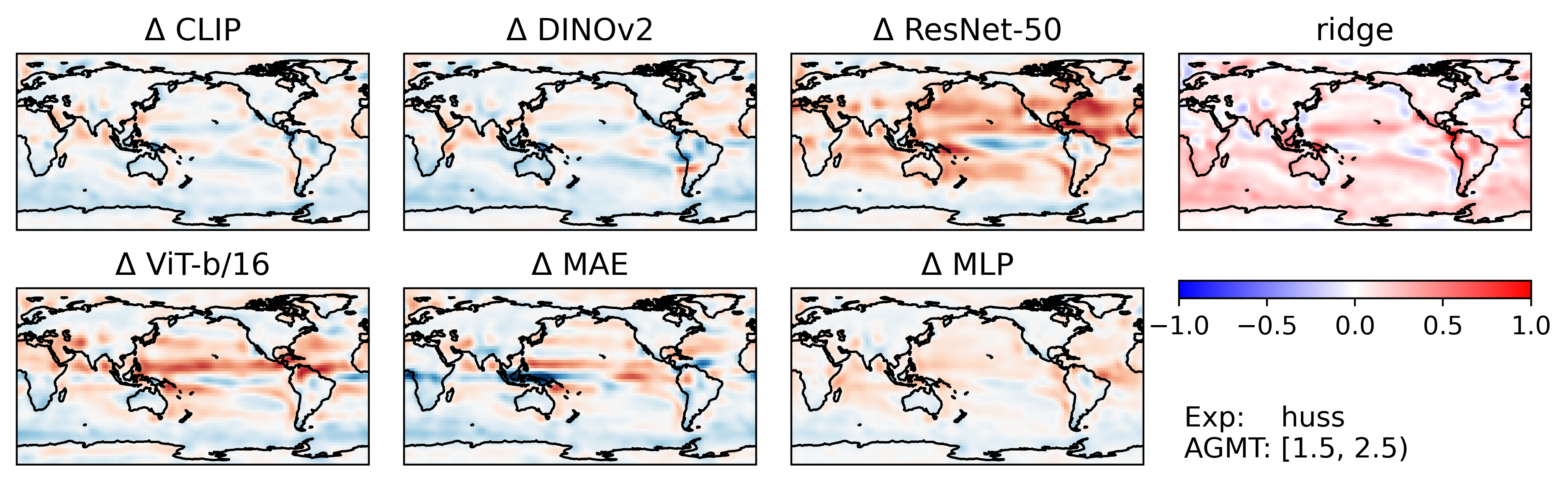}
    \caption{Similar to Figure \ref{fig:IG}, except for the \textbf{huss} experiment}
    \label{fig:si_IG_huss}
\end{figure*}

\begin{figure*}[h]
    \centering
    \includegraphics[width=.9\linewidth]{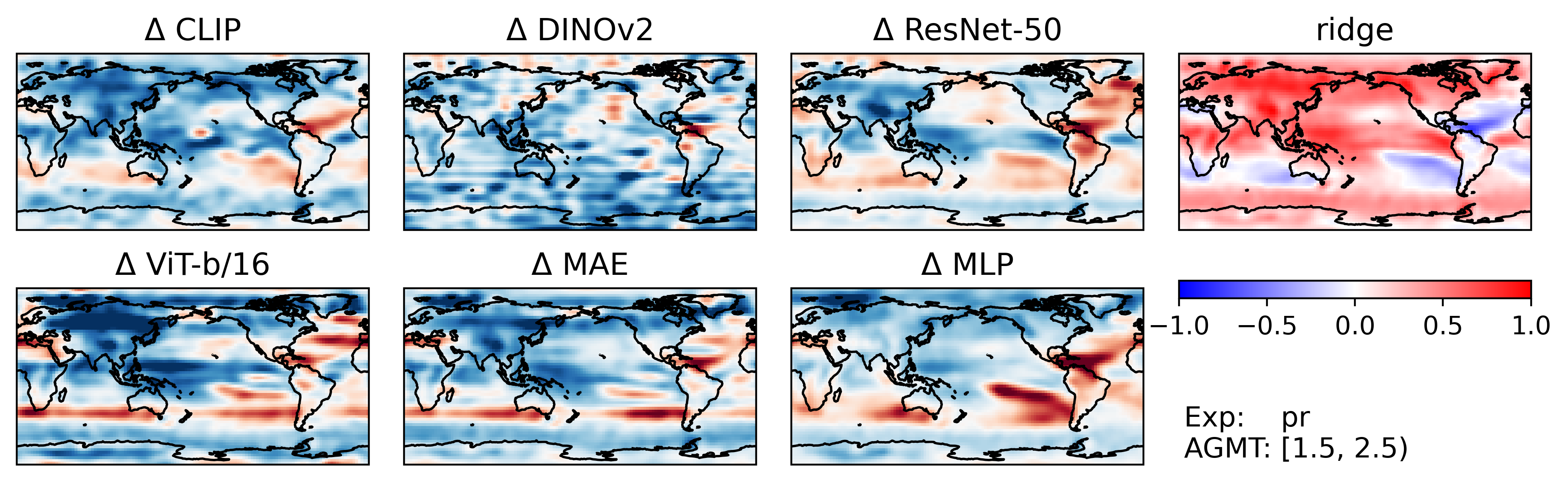}
    \caption{Similar to Figure \ref{fig:IG}, except for the \textbf{pr} experiment}
    \label{fig:si_IG_pr}
\end{figure*}

\begin{figure*}[h]
    \centering
    \includegraphics[width=.9\linewidth]{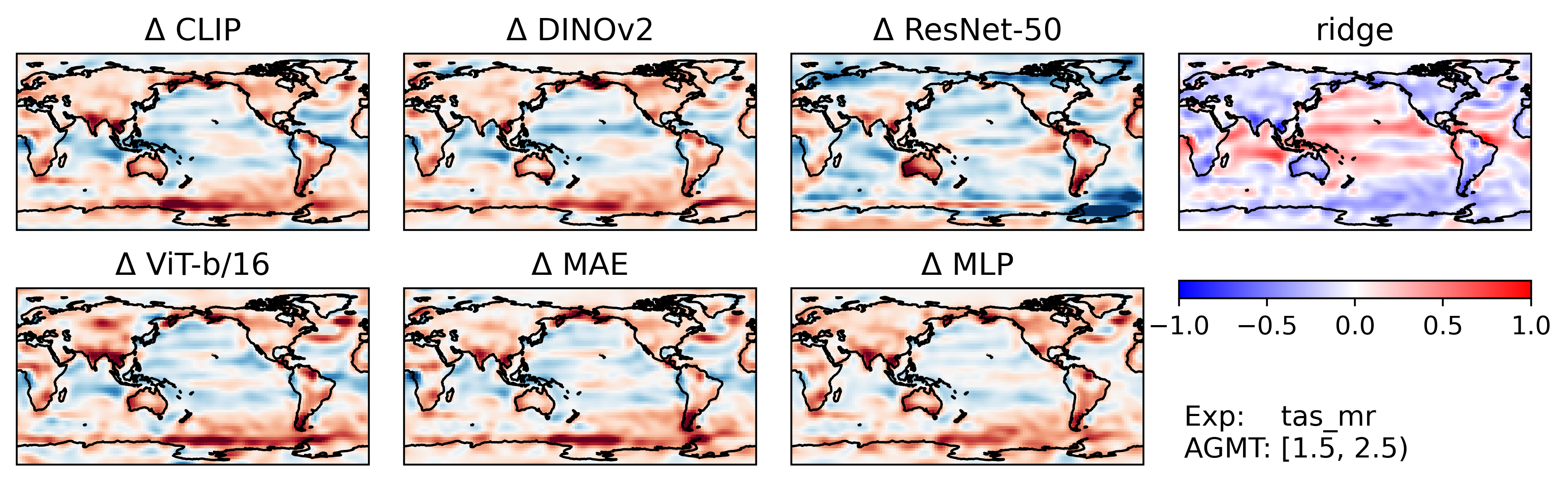}
    \caption{Similar to Figure \ref{fig:IG}, except for the \textbf{tas\_mr} experiment}
    \label{fig:si_IG_tas_mr}
\end{figure*}

\clearpage
\section{CMIP6 Model Information}
\label{App:cmip6_info}

\begin{table*}[h]
    \centering
\caption{Details of CMIP6 simulations incorporated in the training set}
\begin{tabular}{llll}
\toprule
model & scenario &  grid & ensemble members \\
\midrule
\multirow[t]{3}{*}{CESM2} & historical & gn & r1i1p1f1, r4i1p1f1, r11i1p1f1 \\
 & ssp245 & gn & r4i1p1f1, r10i1p1f1, r11i1p1f1 \\
 & ssp370 & gn & r4i1p1f1, r10i1p1f1, r11i1p1f1 \\
\cline{1-4}
\multirow[t]{3}{*}{CNRM-CM6-1} & historical & gr & r1i1p1f2 \\
 & ssp245 & gr & r1i1p1f2 \\
 & ssp370 & gr & r1i1p1f2 \\
\cline{1-4}
\multirow[t]{3}{*}{CNRM-ESM2-1} & historical & gr & r1i1p1f2 \\
 & ssp245 & gr & r1i1p1f2 \\
 & ssp370 & gr & r1i1p1f2 \\
\cline{1-4}
\multirow[t]{3}{*}{CanESM5} & historical & gn & r1i1p1f1, r2i1p1f1, r3i1p1f1, r1i1p2f1, r2i1p2f1, r3i1p2f1 \\
 & ssp245 & gn & r1i1p1f1, r2i1p1f1, r3i1p1f1, r1i1p2f1, r2i1p2f1, r3i1p2f1 \\
 & ssp370 & gn & r1i1p1f1, r2i1p1f1, r3i1p1f1, r1i1p2f1, r2i1p2f1, r3i1p2f1 \\
\cline{1-4}
\multirow[t]{3}{*}{EC-Earth3} & historical & gr & r1i1p1f1, r4i1p1f1, r7i1p1f1 \\
 & ssp245 & gr & r1i1p1f1, r4i1p1f1, r7i1p1f1 \\
 & ssp370 & gr & r1i1p1f1, r4i1p1f1, r150i1p1f1 \\
\cline{1-4}
\multirow[t]{2}{*}{EC-Earth3-CC} & historical & gr & r1i1p1f1 \\
 & ssp245 & gr & r1i1p1f1 \\
\cline{1-4}
\multirow[t]{3}{*}{EC-Earth3-Veg} & historical & gr & r1i1p1f1, r4i1p1f1 \\
 & ssp245 & gr & r1i1p1f1, r4i1p1f1, r6i1p1f1 \\
 & ssp370 & gr & r1i1p1f1, r4i1p1f1 \\
\cline{1-4}
\multirow[t]{3}{*}{EC-Earth3-Veg-LR} & historical & gr & r1i1p1f1, r2i1p1f1, r3i1p1f1 \\
 & ssp245 & gr & r1i1p1f1, r2i1p1f1, r3i1p1f1 \\
 & ssp370 & gr & r1i1p1f1, r2i1p1f1, r3i1p1f1 \\
\cline{1-4}
\multirow[t]{2}{*}{HadGEM3-GC31-LL} & historical & gn & r1i1p1f3 \\
 & ssp245 & gn & r1i1p1f3 \\
\cline{1-4}
\multirow[t]{3}{*}{INM-CM4-8} & historical & gr1 & r1i1p1f1 \\
 & ssp245 & gr1 & r1i1p1f1 \\
 & ssp370 & gr1 & r1i1p1f1 \\
\cline{1-4}
\multirow[t]{3}{*}{INM-CM5-0} & historical & gr1 & r1i1p1f1, r2i1p1f1, r3i1p1f1 \\
 & ssp245 & gr1 & r1i1p1f1 \\
 & ssp370 & gr1 & r1i1p1f1, r2i1p1f1, r3i1p1f1, r4i1p1f1, r5i1p1f1 \\
\cline{1-4}
\multirow[t]{2}{*}{MIROC-ES2L} & historical & gn & r1i1p1f2 \\
 & ssp245 & gn & r1i1p1f2 \\
\cline{1-4}
\multirow[t]{3}{*}{MPI-ESM1-2-HR} & historical & gn & r1i1p1f1, r2i1p1f1, r3i1p1f1 \\
 & ssp245 & gn & r1i1p1f1, r2i1p1f1 \\
 & ssp370 & gn & r1i1p1f1, r2i1p1f1, r3i1p1f1, r4i1p1f1 \\
\cline{1-4}
\multirow[t]{3}{*}{MPI-ESM1-2-LR} & historical & gn & r1i1p1f1, r2i1p1f1, r3i1p1f1 \\
 & ssp245 & gn & r1i1p1f1, r2i1p1f1 \\
 & ssp370 & gn & r1i1p1f1, r2i1p1f1, r3i1p1f1, r4i1p1f1 \\
\cline{1-4}
\multirow[t]{3}{*}{NorESM2-LM} & historical & gn & r1i1p1f1, r2i1p1f1 \\
 & ssp245 & gn & r1i1p1f1, r2i1p1f1, r3i1p1f1 \\
 & ssp370 & gn & r1i1p1f1 \\
\cline{1-4}
\multirow[t]{3}{*}{UKESM1-0-LL} & historical & gn & r1i1p1f2, r2i1p1f2, r3i1p1f2 \\
 & ssp245 & gn & r1i1p1f2, r2i1p1f2, r3i1p1f2 \\
 & ssp370 & gn & r1i1p1f2, r2i1p1f2, r3i1p1f2 \\
\cline{1-4}
\bottomrule
\end{tabular}
\end{table*}

\begin{table*}[h]
    \centering    
\caption{Details of CMIP6 simulations incorporated in the validation set}
\begin{tabular}{llll}
\toprule
model & scenario &  grid & ensemble members \\
\midrule
\multirow[t]{3}{*}{CESM2-WACCM} & historical & gn & r1i1p1f1 \\
 & ssp245 & gn & r1i1p1f1 \\
 & ssp370 & gn & r1i1p1f1 \\
\cline{1-4}
\multirow[t]{3}{*}{KACE-1-0-G} & historical & gr & r1i1p1f1 \\
 & ssp245 & gr & r1i1p1f1 \\
 & ssp370 & gr & r1i1p1f1 \\
\cline{1-4}
\multirow[t]{3}{*}{MRI-ESM2-0} & historical & gn & r1i1p1f1 \\
 & ssp245 & gn & r1i1p1f1 \\
 & ssp370 & gn & r1i1p1f1 \\
\cline{1-4}
\multirow[t]{3}{*}{NorESM2-MM} & historical & gn & r1i1p1f1 \\
 & ssp245 & gn & r1i1p1f1 \\
 & ssp370 & gn & r1i1p1f1 \\
\cline{1-4}
\multirow[t]{2}{*}{TaiESM1} & historical & gn & r1i1p1f1 \\
 & ssp245 & gn & r1i1p1f1 \\
\cline{1-4}
\bottomrule
\end{tabular}
\end{table*}

\begin{table*}[h]
    \centering
\caption{Details of CMIP6 simulations incorporated in the test set}
\begin{tabular}{llll}
\toprule
model & scenario &  grid & ensemble members \\
\midrule
\multirow[t]{3}{*}{ACCESS-CM2} & historical & gn & r1i1p1f1 \\
 & ssp245 & gn & r1i1p1f1 \\
 & ssp370 & gn & r1i1p1f1 \\
\cline{1-4}
\multirow[t]{3}{*}{CMCC-CM2-SR5} & historical & gn & r1i1p1f1 \\
 & ssp245 & gn & r1i1p1f1 \\
 & ssp370 & gn & r1i1p1f1 \\
\cline{1-4}
\multirow[t]{3}{*}{CMCC-ESM2} & historical & gn & r1i1p1f1 \\
 & ssp245 & gn & r1i1p1f1 \\
 & ssp370 & gn & r1i1p1f1 \\
\cline{1-4}
\multirow[t]{2}{*}{FGOALS-g3} & historical & gn & r1i1p1f1 \\
 & ssp245 & gn & r1i1p1f1 \\
\cline{1-4}
\multirow[t]{2}{*}{GFDL-CM4} & historical & gr1 & r1i1p1f1 \\
 & ssp245 & gr1, gr2 & r1i1p1f1 \\
\cline{1-4}
\multirow[t]{3}{*}{GFDL-ESM4} & historical & gr1 & r1i1p1f1 \\
 & ssp245 & gr1 & r1i1p1f1 \\
 & ssp370 & gr1 & r1i1p1f1 \\
\cline{1-4}
\multirow[t]{2}{*}{IITM-ESM} & historical & gn & r1i1p1f1 \\
 & ssp370 & gn & r1i1p1f1 \\
\cline{1-4}
\bottomrule
\end{tabular}
\end{table*}

\end{document}